\documentclass[runningheads]{llncs}
\usepackage{graphicx}
\usepackage[utf8]{inputenc}
\usepackage[T1]{fontenc}
\usepackage{cite}
\usepackage{booktabs}
\usepackage{multirow}
\usepackage{amsmath}
\PassOptionsToPackage{hyphens}{url}\usepackage[hidelinks]{hyperref}
\usepackage{placeins}

\begin{document}
\title{Generating abstractive summaries of Lithuanian news articles using a transformer model}
\titlerunning{Generating abstractive summaries of Lithuanian news articles}
\author{Lukas Stankevičius\inst{1}\orcidID{0000-0003-0012-5471} \and \\
Mantas Lukoševičius\inst{1}\orcidID{0000-0001-7963-285X}}
\authorrunning{L. Stankevičius and M. Lukoševičius.}

\institute{Faculty of Informatics, Kaunas University of Technology, Kaunas, Lithuania}
\maketitle              %
\begin{abstract}
 
In this work, we train the first monolingual Lithuanian transformer model on a relatively large corpus of Lithuanian news articles and compare various output decoding algorithms for abstractive news summarization. We achieve an average ROUGE-2 score 0.163, generated summaries are coherent and look impressive at first glance. However, some of them contain misleading information that is not so easy to spot. We describe all the technical details and share our trained model and accompanying code in an online open-source repository, as well as some characteristic samples of the generated summaries.

\end{abstract}
\section{Introduction}
Recent innovations in deep learning models gave rise to new capabilities for natural language generation. Huge GPT-2\cite{radford2019language} and following GPT-3\cite{brown2020language} models attracted attention of the public with convincing stories from ``new species of unicorns'' to ``colony of humans living in elevator shafts''.\footnote{\url{https://thenextweb.com/neural/2020/10/07/someone-let-a-gpt-3-bot-loose-on-reddit-it-didnt-end-well/}} Even generation of realistic images from a written text prompt is now possible.\footnote{\url{https://openai.com/blog/dall-e/}}

Lithuanian language can not enjoy the same attention and application of innovations as English. There are only sparse records of text generation\footnote{\url{https://www.15min.lt/mokslasit/straipsnis/technologijos/lietuviskas-d-i-ne-dirbtinis-intelektas-o-dirbtinis-idiotas-646-748590} (in Lithuanian)}$^{,}$\footnote{\url{https://ktu.edu/news/ktu-profesorius-saulius-keturakis-nuo-siol-lietuviu-literatura-kuria-ir-masinos/} (in Lithuanian)} which were more of creative projects with no scientific or technical details made public.

One practical application of text generation is summary writing. Now more than ever, growing amounts of information require thorough digestion and simplification. Currently, only extractive text summarization based on classical semantic analysis framework \cite{vitkute2016nlp} is available for Lithuanian texts.\footnote{\url{https://semantika.lt/Analysis/Summary}}

In this work we: (1) train the first Lithuanian monolingual transformer models on news articles summarization task; (2) explore different decoding options for the best summary generation; (3) share all the technical details and code for reproduction. We hope that this work will give a boost to the Lithuanian language generation as well as useful experience transferable to other mid-resource languages.

\section{Previous Related Work}

\subsection{From $n$-gram to Transformer Language Models}

Before the success of neural language models, statistical ones were the most common. One example is the $n$-gram language model. For chosen $n$, it approximates the conditional probability of the next word $w_i$, given a word sequence of length $i-1$ behind, by
\begin{equation}
  P(w_i|w_{1:i-1}) \approx P(w_i|w_{i-n:i-1}).
\end{equation}
In practice, this means collecting $n$-gram counts from the corpus and normalizing them to probabilities. The higher $n$-gram is used, the bigger improvements are expected. However, increasing history size is difficult as it requires a very large corpus to produce sufficient counts and suffers from sparsity. It also leads to computational overhead, requires a lot of memory.

Neural language models offer several advantages over statistical ones. Instead of using predictive distributions determined by counting, they use internal representation to perform a high-dimensional interpolation between training examples \cite{graves2013generating}. This way neural networks are better at generating real-valued data than exact matches. Finally, these models do not suffer from the curse of dimensionality.

The first neural networks successfully adopted for language modeling were Recurrent Neural Networks (RNNs) of Long Short-Term Memory (LSTM) type \cite{hochreiter1997long, graves2013generating, sundermeyer2012lstm}. They were also one of the first to be employed for abstractive text summarization \cite{rush2015neural}. The recurrent nature of this architecture makes it very simple to feed sequential text of varying length, while enhancements over simpler RNN variants increase stability with the added memory. It was a huge improvement over the $n$-gram language models but had its own drawbacks nonetheless.

Computations with recurrent neural networks do not scale well, because inputs to the model must be passed sequentially. Each word in a sequence must wait for the computation of the hidden state of the previous word to complete. %
This limits the parallelization of the computations.

The other drawback is that information of all the previous inputs has to 
fit into the last hidden state. As a result, only recent inputs are usually sufficiently ``remembered'', and the model is unable to model long-range dependencies between inputs. This was only partially addressed by deeper processing of the hidden states through specific units (LSTM), using Bi-directional models, and employing attention mechanisms \cite{bahdanau2014neural}.

Almost all of the above issues of neural language models were solved by the current state-of-the-art Transformer architecture \cite{vaswani2017attention}. The whole sequence now can be fed to a model at once, while added positional embeddings preserve the order information. Large data and computation resources can now be fully utilized. Additionally, due to the attention mechanism, each input word can attend to any other in any layer, thus producing contextualized word vectors.

As of now, there exist multiple Transformer type models. Many implementations can be found in the Huggingface Transformers library \cite{wolf-etal-2020-transformers}. Although there are efforts \cite{wang2019bert} to use models trained on masked words predictions \cite{devlin2018bert}, usually the ones trained on language modeling (predicting the next word) are best suited for text generation. Specifically for the summarization task, the more notable Transformer models are: T5 \cite{raffel2020exploring}, BART \cite {lewis2020bart}, PEGASUS \cite{zhang2020pegasus}, and ProphetNet \cite{qi2020prophetnet}.

\subsection{Decoding Algorithms} \label{decoding}

Given a dataset $D = \{x^{1}, \ldots, x^{|D|}\}$, current state-of-the-art text generation models \cite{raffel2019exploring} train a neural network with parameters $\theta$ to minimize the negative log-likelihood
\begin{equation}
  \mathcal{L}(D) = -\sum_{k=1}^{|D|}\log\prod_{i=1}^{|x^{k}|} p_{\theta}(x_{i}^{k}|x_{<i}^{k}),
\end{equation}
where $x_{i}^{k}$ is the $i$-th word of the $k$-th text in the dataset $D$.

After this kind of training, candidate word probabilities can be inferred for all words in a vocabulary. The chosen word can later be appended to conditional input to predict the next one. The process repeats for the desired output sequence length or until the special token (i.e., <eos> for T5) is reached. Also, additional words can be added to condition the model more specifically \cite{keskar2019ctrl, raffel2020exploring}.

Below we will describe various decoding algorithms for choosing the next word from given probabilities of all the words in a vocabulary.

\subsubsection{Maximization-Based Decoding}

These are greedy, deterministic decoding methods. They assume that the model assigns higher probabilities for higher-quality text.

The most simple decoding is called \textbf{greedy search}. It selects the next word as the one with the highest probability. This simplicity makes it very fast, albeit not optimal.

Another popular, heuristic expanding the one above, is \textbf{beam search} decoding. Instead of choosing only one word with the highest probability at a time, a defined number of word sequences with the highest overall probabilities are kept. This way, a single low-probability word would not shadow a high-overall-probability sequence.

A natural way to improve beam search is to use a higher beam size. An obvious drawback is that the computation intensifies. A second disadvantage, as noticed in \cite{tu2017neural, koehn2017six} for translation tasks, is that increasing beam size reduces BLEU score (mentioned in Section \ref{other_evaluation}). This is explained by higher beam sizes generating shorter sequences. This effect is especially strong when beam sizes are in the range of 100-1\,000. According to \cite{nlp_gen_lecture}, low beam sizes produce more on-topic but nonsensical text, while the high ones converge to a correct and generic, but less relevant response.

\subsubsection{Sampling}

As argued in \cite{holtzman2019curious}, human language does not follow a distribution of high probability next words, in contrast to what greedy decoding methods do. This issue is alleviated by stochastic decoding methods.

In the most basic form, sampling is used to randomly select the next word according to its conditional probability distribution. Instead of sampling from the distribution of all vocabulary words, \textbf{Top-$K$} sampling was proposed in \cite{fan2018hierarchical}. Here $K$ most probable next words are selected and the probability mass is redistributed among only those words. 

Later \textbf{Top-$p$} (nucleus) sampling \cite{holtzman2019curious} was suggested to adapt to different initial probability distributions. Here the most probable words are selected such, that the sum of their probabilities is not greater than $p$. This way an adaptation to sharp or flat probability distributions is implemented.

\subsection{Additional Techniques}

During language modeling, the last neural network layer produces output for each word in vocabulary. These outputs are raw values called ``logits''. To convert them into probabilities, commonly a softmax function is used:
\begin{equation}
  \text{softmax}(y_i) = \frac{\exp(y_{i})}{\sum_{j}\exp(y_{j})}.
\end{equation}

One technique to rebalance the probabilities is a softmax temperature parameter $\tau$. Then the probability of word $i$ given all vocabulary logits $\mathbf{y}$ is 
\begin{equation}
  P_{i} = \text{softmax}(y_{i}/\tau) = \frac{\exp{(y_{i}/\tau)}}{\sum_{j}\exp{(y_{j}/\tau)}}.
\end{equation}
Then $\tau>1$, the probability distribution becomes more uniform, and more diverse outputs are expected. Otherwise, the probability concentrates on a smaller number of top words.

\subsection{Evaluation Methods}

It is very difficult to evaluate the quality of abstractive summarization. Even if one has reference summaries, alternative correct summaries can be easily produced using different words, their order, sentence length, emphasis, etc. The most accurate evaluation thus would be done by humans. Yet it is very expensive and slow, conducting it effectively is difficult. Due to these reasons, automatic evaluation metrics are widely used.

\subsubsection{ROUGE}

Introduced by \cite{lin2004rouge} and abbreviated from Recall-Oriented Understudy for Gisting Evaluation (ROUGE), they are word-overlap-based metrics. The following ROUGE metrics are regularly used for summary evaluation.

\paragraph{ROUGE-$n$}
Let $p$ be ``the number of common $n$-grams between the candidate and the reference summaries'', and $q$ be ``the number of $n$-grams extracted from the reference summary only'' \cite{allahyari2017text}. Then the recall score is computed as 
\begin{equation}
  \text{ROUGE-}n =\frac{p}{q}.
\end{equation}
The precision score is divided by ``the number of $n$-grams extracted from the generated summary only'' instead, and F-score combines the precision and the recall. We report F-scores in our results. Typically, $n$ values of 1 and 2 are used.%

\paragraph{ROUGE-$L$}

This metric calculates the length of the longest common subsequence between the candidate and the reference summaries.

For example, on English news dataset CNN / Daily Mail \cite{hermann2015teaching, nallapati2016abstractive}, the highest ROUGE scores using abstractive summarization currently are reported at ROUGE-1 = 45.94, ROUGE-2 = 22.32, and ROUGE-$L$ = 42.48 \cite{dou2020gsum}.

\subsubsection{Other Evaluation Metrics}\label{other_evaluation}

One of the attempts to capture the overlap between meanings instead of exact words is to compare word vectors. BERTScore \cite{zhang2019bertscore} is a model-based metric that uses a pretrained BERT \cite{devlin2018bert} model to produce contextualized embeddings and matches words in the candidate and reference sentences by cosine similarity. Compared to ROUGE, however, this evaluation method is more compute-intensive and lacks the simple explainability.

In machine translation, BLEU \cite{papineni2002bleu} is considered the standard evaluation metric. BLEU is based on precision, while ROUGE can be based more on recall and thus is more suitable for the summarization task.

\section{Data}\label{data}

We crawled news articles with summary and the main text (body) parts from the most popular Lithuanian news websites. We filtered data so that summary > 10 and the main text > 100 characters in length. As the goal of summarization is to produce a shorter text than the original, we used only the articles where the main text length in characters was at least twice the summary length. 
\begin{table}
\centering
\caption{Our data corpus statistics}
\label{table:data_info}
\setlength\tabcolsep{4pt}
\begin{tabular}{lrcc}
\toprule
\multirow{2.5}{*}{Website} &  \multirow{2.5}{*}{Article count}   & \multicolumn{2}{c} {Time period} \\
\cmidrule(l){3-4}
       &         &         From       &  To       \\
\midrule
15min.lt         &  767\,182 &     2007-07-09 &    2020-09-23 \\
lrytas.lt        &  522\,519 &     2017-03-30 &    2020-09-24 \\
delfi.lt         &  436\,487 &     2000-01-27 &    2020-09-25 \\
lrt.lt           &  181\,356 &     2012-05-25 &    2020-09-24 \\
technologijos.lt &   66\,867 &     2007-02-22 &    2020-09-24 \\
bernardinai.lt   &   25\,250 &     2004-03-30 &    2020-04-29 \\
kasvyksta.lt     &   18\,781 &     2012-03-15 &    2020-09-25 \\
panele.lt        &   11\,587 &     2007-03-07 &    2020-09-24 \\
kaunodiena.lt    &    1\,485 &     2002-08-05 &    2014-06-25 \\
\bottomrule
\end{tabular}
\end{table}

We want our models to learn abstractive summarization and avoid copying. Due to this reason, for each summary and main text pair, we found the longest matching string sequence and calculated the overlap as a ratio of this sequence and the summary lengths. We left only pairs with this overlap ratio less than 0.2. This criterion is similar to summaries not having very high ROUGE-$L$ scores compared to the main texts.

The final filtered dataset consists of 2\,031\,514 news articles. Detailed statistics are depicted in Table \ref{table:data_info}. We put random 4\,096 articles from this set aside from training for validation.

\section{Methods}

We used SentencePiece \cite{kudo2018sentencepiece} to encode text
as unigram \cite{kudo2018subword} subwords. A tokenizer was trained on $10^{6}$ samples of our text corpus and was set to have a vocabulary size of 32\,000 (the same size as the English tokenizer from \cite{raffel2020exploring}).

We used T5 base \cite{raffel2020exploring} transformer model implementation from \cite{wolf-etal-2020-transformers}. This library also contains methods we used in this work for text generation. The model was trained for 350\,000 steps with batches of 128 text-summary pairs (achieved by 32 gradient accumulation steps), both truncated to 512 tokens. Using mixed precision and GeForce RTX 2080 Ti GPU it took approximately 500 hours and consisted of 22 passes through the dataset (epochs). We used Adafactor \cite{shazeer2018adafactor} optimizer with 10\,000 warm-up steps followed by inverse square root internal learning rate schedule.

We initialized the weights with a pretrained English \textit{t5-base} model, as it showed a faster convergence compared to random weight initialization (see Figure \ref{fig2}). We do not use a pretrained multilingual \textit{mt5-base} model \cite{xue2020mt5} because: (1) due to shorter tokens, tokenized sequences are on average 1.49 times longer and (2) it has 580\,M parameters versus our used 220\,M mainly due to the bigger multilingual vocabulary embedding matrix. These reasons made training multilingual \textit{mt5-base} 4 times slower than a monolingual model based on \textit{t5-base} and higher ROUGE scores were faster reached with the latter.

We decided that the generated sample is repetitive if any constituent word count, except the stop word ``ir'' (``and'' in Lithuanian), is greater than 7. We also calculated a generated text length fraction as a ratio of the generated text to the target summary character counts. ROUGE scores were calculated for stemmed texts.

\section{Results}

In this section, we first report results with a model trained for 65\,000 steps and then see the changes continuing the training further.

\subsection{Repetition}

We found out that the best way to avoid repetition is to obstruct the repeated generation of 2-grams by setting\texttt{ no\_repeat\_ngram\_size }parameter value to 2. This way 36 repetitive samples (out of 4\,096) of greedy beam search with beam size 10 were reduced to 0.

\subsection{Reshaping Probability Distribution}

To our surprise, greedy methods gave quite decent results. Any attempt to reshape probability distribution favored tokens with the top probabilities. For sampling, we tried temperatures $\tau$ values 0.8, 0.6, 0.4, 0.2, 0.1, and 0.05, with the last one yielding the best scores, ROUGE-2 = 0.132. We also experimented with Top-$p$ sampling trying $p$ values of 0.9, 0.8, 0.6, 0.4, 0.2, and 0.1, with the last one also yielding the best scores, ROUGE-2 = 0.131. Both decoding methods had all metrics approximately the same as greedy search. Top-$k$ similarly resulted in best $k=5$ (tried 50, 40, 30, 20, 10, 5), ROUGE-2 = 0.109.

\subsection{Best Decoding Methods}

The best results were obtained with beam search decoding (see Table \ref{table:results_65}). There was no significant difference between greedy and Top-$k$ ($k$=50) beam searches. Though the latter is expected to be more diverse due to its stochastic nature. Increasing beam size from 10 to 20 only increased the generation time to more than 3 hours and did not benefit ROUGE scores.

\begin{figure}
\centerline{\includegraphics[scale=0.67]{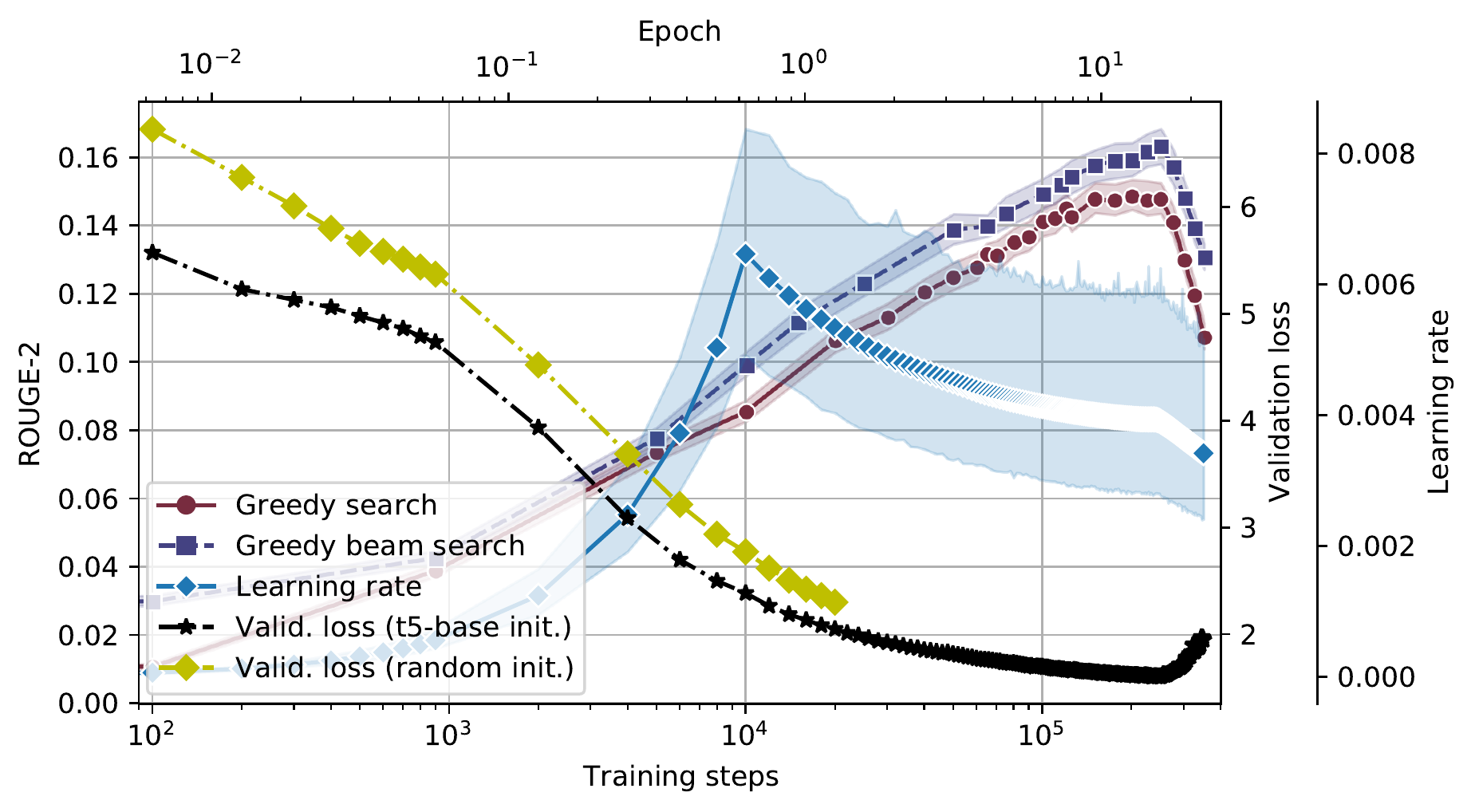}}
\caption{ROUGE-2 mean F-score, learning rate, and validation loss dependency on model training steps. Shadow areas correspond to confidence interval of 95\,\%. ROUGE-2 and learning rate are shown for model trained with \textit{t5-base} initialization. }
\label{fig2}
\centering
\end{figure}

Training the model longer is beneficial. After 250\,000 steps, mean ROUGE-2 reached 0.148 for greedy, 0.163 for greedy beam search, and 0.161 for Top-50 beam search decoding (see Figure \ref{fig2}). We use ROUGE-2 as the main metric here, as it seems to be the hardest of the three to get high values. Training even further, we observe overfitting: the validation loss begins to ascent, and ROUGE metrics deteriorate (note, however, the log-scale of the x-axis).

\begin{table}
\centering
\caption{Text generation performance metrics on 4\,096 validation articles of model trained for 65\,000 steps. For ROUGE F-scores are given. All methods use\texttt{ no\_repeat\_ngram\_size} $=2$, ten beams for beam searches.}
\label{table:results_65}
\setlength\tabcolsep{4pt}
\begin{tabular}{lrrrr}
\toprule

\multirow{2.5}{*}{Decoding method} & \multicolumn{4}{c}{Performance metric: mean (standard deviation)} \\
\cmidrule(l){2-5}
       &  ROUGE-1 & ROUGE-2 & ROUGE-$L$  & Length fraction   \\
\midrule
Greedy search      &  0.298 (0.154) &  0.132 (0.137) & 0.233 (0.147) & 0.79 (0.40) \\
Greedy beam search &  0.303 (0.162) &  0.140 (0.146) & 0.238 (0.155) & 0.82 (0.39) \\
Top-50 beam search &  0.306 (0.156) &  0.138 (0.143) & 0.235 (0.152) & 1.07 (0.67)  \\

\bottomrule
\end{tabular}
\end{table}

We discuss the qualitative analysis of the generated summaries in the next Section \ref{discussion} and provide some illustrative generated summaries in Appendix \ref{appendix}.

\section{Conclusion and Discussion of Results}\label{discussion}

We trained the first reported monolingual Lithuanian transformer model and compared various decoding algorithms for abstractive news summarization. We used a moderately-sized modern universal transformer model T5, because of the computational requirements as well as the availability of training texts. We still observe that the model tends to overfit if trained for too long.

The best configuration we achieved is the model trained for 250\,000 steps and greedy beam decoding with 10 beams. It achieved ROUGE-2 = 0.163. 

Mostly all the generated summaries were coherent and made the first impression of being professionally written. Some of them successfully collected important information from various parts of the input text and summarized the main idea of the text quite impressively. 

However, a closer examination of the generated samples reveals some chaotic and misleading summaries. There were a lot of text samples with minor factual mistakes, such as a mistaken number, abbreviation, name, day of the week, that can often be identified only by reading the main body text carefully or comparing with the target summary. Sometimes irrelevant information was added to the generated summaries, likely the one too often seen during the training. An example dynamic of summaries generated of a single text during the training is depicted in Appendix \ref{A1}. While all the summaries above 0 steps sound convincing, completely factually correct summaries are only at 175\,000 and 300\,000 steps, others have small errors.

A lot of high-ROUGE samples were finely combined multiple excerpts from the main text and learned templates repeating in our target data. On the other hand, there were also very good abstractive summaries with ROUGE scores just above zero. This emphasizes both the need for thorough data preparation and better evaluation of the summary quality. %
The latter could be covered with more resource-intensive scores like BERTScore \cite{zhang2019bertscore} or manual human evaluation.

Overall, our trained model is good for generating fiction but makes trusting the generated news summaries problematic. Thus, for real-world applications, a more substantial human intervention is needed for quality assessment/control. 

We hope that this work will boost adopting transformers for the Lithuanian language. We make our trained model and code available at %
\url{https://github.com/LukasStankevicius/Generating-abstractive-summaries-of-Lithuanian-news-articles-using-a-transformer-model}.

\bibliographystyle{splncs04}
\bibliography{references}

\begin{thebibliography}{10}
\providecommand{\url}[1]{\texttt{#1}}
\providecommand{\urlprefix}{URL }
\providecommand{\doi}[1]{https://doi.org/#1}

\bibitem{allahyari2017text}
Allahyari, M., Pouriyeh, S., Assefi, M., Safaei, S., Trippe, E.D., Gutierrez,
  J.B., Kochut, K.: Text summarization techniques: a brief survey. arXiv
  preprint arXiv:1707.02268  (2017)

\bibitem{bahdanau2014neural}
Bahdanau, D., Cho, K., Bengio, Y.: Neural machine translation by jointly
  learning to align and translate. arXiv preprint arXiv:1409.0473  (2014)

\bibitem{brown2020language}
Brown, T.B., Mann, B., Ryder, N., Subbiah, M., Kaplan, J., Dhariwal, P.,
  Neelakantan, A., Shyam, P., Sastry, G., Askell, A., et~al.: Language models
  are few-shot learners. arXiv preprint arXiv:2005.14165  (2020)

\bibitem{devlin2018bert}
Devlin, J., Chang, M.W., Lee, K., Toutanova, K.: Bert: Pre-training of deep
  bidirectional transformers for language understanding. arXiv preprint
  arXiv:1810.04805  (2018)

\bibitem{dou2020gsum}
Dou, Z.Y., Liu, P., Hayashi, H., Jiang, Z., Neubig, G.: {GS}um: A general
  framework for guided neural abstractive summarization. arXiv preprint
  arXiv:2010.08014  (2020)

\bibitem{fan2018hierarchical}
Fan, A., Lewis, M., Dauphin, Y.: Hierarchical neural story generation. In:
  Proceedings of the 56th Annual Meeting of the Association for Computational
  Linguistics (Volume 1: Long Papers). pp. 889--898 (2018)

\bibitem{graves2013generating}
Graves, A.: Generating sequences with recurrent neural networks. arXiv preprint
  arXiv:1308.0850  (2013)

\bibitem{hermann2015teaching}
Hermann, K.M., Ko{\v{c}}isk{\`y}, T., Grefenstette, E., Espeholt, L., Kay, W.,
  Suleyman, M., Blunsom, P.: Teaching machines to read and comprehend. arXiv
  preprint arXiv:1506.03340  (2015)

\bibitem{hochreiter1997long}
Hochreiter, S., Schmidhuber, J.: Long short-term memory. Neural computation
  \textbf{9}(8),  1735--1780 (1997)

\bibitem{holtzman2019curious}
Holtzman, A., Buys, J., Du, L., Forbes, M., Choi, Y.: The curious case of
  neural text degeneration. In: International Conference on Learning
  Representations (2019)

\bibitem{keskar2019ctrl}
Keskar, N.S., McCann, B., Varshney, L.R., Xiong, C., Socher, R.: Ctrl: A
  conditional transformer language model for controllable generation. arXiv
  preprint arXiv:1909.05858  (2019)

\bibitem{koehn2017six}
Koehn, P., Knowles, R.: Six challenges for neural machine translation. arXiv
  preprint arXiv:1706.03872  (2017)

\bibitem{kudo2018subword}
Kudo, T.: Subword regularization: Improving neural network translation models
  with multiple subword candidates. In: Proceedings of the 56th Annual Meeting
  of the Association for Computational Linguistics (Volume 1: Long Papers). pp.
  66--75 (2018)

\bibitem{kudo2018sentencepiece}
Kudo, T., Richardson, J.: Sentence{P}iece: A simple and language independent
  subword tokenizer and detokenizer for neural text processing. In: Proceedings
  of the 2018 Conference on Empirical Methods in Natural Language Processing:
  System Demonstrations. pp. 66--71 (2018)

\bibitem{lewis2020bart}
Lewis, M., Liu, Y., Goyal, N., Ghazvininejad, M., Mohamed, A., Levy, O.,
  Stoyanov, V., Zettlemoyer, L.: {BART}: Denoising sequence-to-sequence
  pre-training for natural language generation, translation, and comprehension.
  In: Proceedings of the 58th Annual Meeting of the Association for
  Computational Linguistics. pp. 7871--7880 (2020)

\bibitem{lin2004rouge}
Lin, C.Y.: Rouge: A package for automatic evaluation of summaries. In: Text
  summarization branches out. pp. 74--81 (2004)

\bibitem{nlp_gen_lecture}
Manning, C., See, A.: Stanford {CS224N}: {NLP} with deep learning | winter 2019
  | lecture 15 – natural language generation (2019),
  \url{https://youtu.be/4uG1NMKNWCU}

\bibitem{nallapati2016abstractive}
Nallapati, R., Zhou, B., dos Santos, C., G{\.{u}}l{\c{c}}ehre, {\c{C}}., Xiang,
  B.: Abstractive text summarization using sequence-to-sequence {RNN}s and
  beyond. In: Proceedings of The 20th SIGNLL Conference on Computational
  Natural Language Learning. pp. 280--290 (2016)

\bibitem{papineni2002bleu}
Papineni, K., Roukos, S., Ward, T., Zhu, W.J.: Bleu: a method for automatic
  evaluation of machine translation. In: Proceedings of the 40th annual meeting
  of the Association for Computational Linguistics. pp. 311--318 (2002)

\bibitem{qi2020prophetnet}
Qi, W., Yan, Y., Gong, Y., Liu, D., Duan, N., Chen, J., Zhang, R., Zhou, M.:
  Prophet{N}et: Predicting future n-gram for sequence-to-sequence pre-training.
  In: Proceedings of the 2020 Conference on Empirical Methods in Natural
  Language Processing: Findings. pp. 2401--2410 (2020)

\bibitem{radford2019language}
Radford, A., Wu, J., Child, R., Luan, D., Amodei, D., Sutskever, I.: Language
  models are unsupervised multitask learners. OpenAI blog  \textbf{1}(8), ~9
  (2019)

\bibitem{raffel2019exploring}
Raffel, C., Shazeer, N., Roberts, A., Lee, K., Narang, S., Matena, M., Zhou,
  Y., Li, W., Liu, P.J.: Exploring the limits of transfer learning with a
  unified text-to-text transformer. arXiv preprint arXiv:1910.10683  (2019)

\bibitem{raffel2020exploring}
Raffel, C., Shazeer, N., Roberts, A., Lee, K., Narang, S., Matena, M., Zhou,
  Y., Li, W., Liu, P.J.: Exploring the limits of transfer learning with a
  unified text-to-text transformer. Journal of Machine Learning Research
  \textbf{21},  1--67 (2020)

\bibitem{rush2015neural}
Rush, A.M., Chopra, S., Weston, J.: A neural attention model for abstractive
  sentence summarization. In: Proceedings of the 2015 Conference on Empirical
  Methods in Natural Language Processing. pp. 379--389 (2015)

\bibitem{shazeer2018adafactor}
Shazeer, N., Stern, M.: Adafactor: Adaptive learning rates with sublinear
  memory cost. In: International Conference on Machine Learning. pp.
  4596--4604. PMLR (2018)

\bibitem{sundermeyer2012lstm}
Sundermeyer, M., Schl{\"u}ter, R., Ney, H.: {LSTM} neural networks for language
  modeling. In: Thirteenth annual conference of the international speech
  communication association (2012)

\bibitem{tu2017neural}
Tu, Z., Liu, Y., Shang, L., Liu, X., Li, H.: Neural machine translation with
  reconstruction. Proceedings of the AAAI Conference on Artificial Intelligence
   \textbf{31}(1) (Feb 2017),
  \url{https://ojs.aaai.org/index.php/AAAI/article/view/10950}

\bibitem{vaswani2017attention}
Vaswani, A., Shazeer, N., Parmar, N., Uszkoreit, J., Jones, L., Gomez, A.N.,
  Kaiser, L., Polosukhin, I.: Attention is all you need. arXiv preprint
  arXiv:1706.03762  (2017)

\bibitem{vitkute2016nlp}
Vitkut{\.e}-Ad{\v{z}}gauskien{\.e}, D., Utka, A., Amilevi{\v{c}}ius, D.,
  Krilavi{\v{c}}ius, T.: {NLP} infrastructure for the {L}ithuanian language.
  In: LREC 2016: 10th international conference on Language resources and
  evaluation, 23-28 May, 2016, Portoro{\v{z}}, Slovenia: proceedings. Paris:
  European Language Resources Association, 2016 (2016)

\bibitem{wang2019bert}
Wang, A., Cho, K.: Bert has a mouth, and it must speak: Bert as a markov random
  field language model. arXiv preprint arXiv:1902.04094  (2019)

\bibitem{wolf-etal-2020-transformers}
Wolf, T., Debut, L., Sanh, V., Chaumond, J., Delangue, C., Moi, A., Cistac, P.,
  Rault, T., Louf, R., Funtowicz, M., Davison, J., Shleifer, S., von Platen,
  P., Ma, C., Jernite, Y., Plu, J., Xu, C., Scao, T.L., Gugger, S., Drame, M.,
  Lhoest, Q., Rush, A.M.: Transformers: State-of-the-art natural language
  processing. In: Proceedings of the 2020 Conference on Empirical Methods in
  Natural Language Processing: System Demonstrations. pp. 38--45. Association
  for Computational Linguistics, Online (Oct 2020),
  \url{https://www.aclweb.org/anthology/2020.emnlp-demos.6}

\bibitem{xue2020mt5}
Xue, L., Constant, N., Roberts, A., Kale, M., Al-Rfou, R., Siddhant, A., Barua,
  A., Raffel, C.: {mT5}: A massively multilingual pre-trained text-to-text
  transformer (2020)

\bibitem{zhang2020pegasus}
Zhang, J., Zhao, Y., Saleh, M., Liu, P.: Pegasus: Pre-training with extracted
  gap-sentences for abstractive summarization. In: International Conference on
  Machine Learning. pp. 11328--11339. PMLR (2020)

\bibitem{zhang2019bertscore}
Zhang, T., Kishore, V., Wu, F., Weinberger, K.Q., Artzi, Y.: {BERTScore}:
  Evaluating text generation with {BERT}. arXiv preprint arXiv:1904.09675
  (2019)

\end{thebibliography}
\appendix

\section{Text sample generation dynamics}\label{appendix}
\subsection{Reference summary} 
Kai Lietuva dar buvo okupuota ir mūsų šalies krepšininkai privalėjo žaisti TSRS rinktinėje, keli jų buvo ryškūs lyderiai.\footnote{The original article and summary are available at \url{https://www.15min.lt/24sek/naujiena/lietuva/tarp-penkiu-rezultatyviausiu-tsrs-rinktines-visu-laiku-zaideju-trys-lietuviai-875-1380030} (in Lithuanian)}
\subsection{Generated summaries}\label{A1}

\paragraph{0 steps, R-1=0.136, R-2=0, R-L=0.091}
Lietuvos krepšinio legenda pelnė po 13,6 taško per 84 mačus. Dešimtuke taip pat yra, ir
\paragraph{25\,000 steps, R-1=0.182, R-2=0, R-L=0.136}
Lietuvos krepšinio federacijos (LKF) sudarytame geriausių visų laikų rezultatyviausių krepšininkų sąraše – Arvydas Sabonis ir Modestas Paulauskas.
\paragraph{75\,000 steps, R-1=0.130, R-2=0.022, R-L=0.109}
Legendinis Lietuvos krepšininkas Arvydas Sabonis pirmadienį socialiniame tinkle pasidalijo statistika, kuri viršūnėje yra Arvydo Sabonio. 1982 metais debiutavęs 222 cm ūgio vidurio puolėjas su raudona apranga sužaidė 52 rungtynes, per kurias rinko po 15,6 taško. Tai pats aukščiausias rezultatyvumo vidurkis tarp visų sovietų komandai atstovavusių žaidėjų, skaičiuojant tuos, kurie sužaidė ne mažiau nei 50 rungtynių.
\paragraph{100\,000 steps, R-1=0.169, R-2=0.049, R-L=0.145}
Lietuvos krepšinio legenda Arvydas Sabonis pirmadienį socialiniame tinkle pasidalijo statistika, kurios viršūnėje yra A.Sabonis. 1982 metais TSRS rinktinėje debiutavęs 222 cm ūgio vidurio puolėjas su raudona apranga sužaidė 52 rungtynes, per kurias rinko po 15,6 taško. Tai pats aukščiausias rezultatyvumo vidurkis tarp visų SSRS komandai atstovavusių žaidėjų.
\paragraph{125\,000 steps, R-1=0.360, R-2=0.083, R-L=0.320}
Lietuvos krepšinio federacijos (LKF) prezidentas Arvydas Sabonis ir buvęs Kauno „Žalgirio“ krepšininkas Rimas Kurtinaitis tapo rezultatyviausiais visų laikų TSRS rinktinės žaidėjais.
\paragraph{150\,000 steps, R-1=0.318, R-2=0.095, R-L=0.227}
Arvydui Saboniui ir Šarūnui Marčiulioniui – rezultatyviausi visų laikų Tėvynės sąjungos (TSRS) rinktinės krepšininkai.
\paragraph{175\,000 steps, R-1=0.207, R-2=0.036, R-L=0.172}
Lietuvos krepšinio federacijos (LKF) prezidento Arvydo Sabonio rezultatyvumo vidurkis yra aukščiausias tarp visų Sovietų Sąjungos rinktinėje atstovavusių žaidėjų, skaičiuojant tuos, kurie sužaidė bent po 50 oficialių rungtynių.
\paragraph{200\,000 steps, R-1=0.197, R-2=0, R-L=0.131}
Legendinis Arvydas Sabonis yra vienas geriausių visų laikų Lietuvos rinktinės krepšininkų, sužaidęs bent po 50 oficialių rungtynių nacionalinėje komandoje. Jis yra pirmas pagal rezultatyvumo vidurkį tarp visų sovietų komandoje atstovavusių krepšininkų.
\paragraph{225\,000 steps, R-1=0.296, R-2=0.077, R-L=0.185}
Legendinis Lietuvos krepšininkas Arvydas Sabonis yra pirmas pagal rezultatyvumą tarp visų Sovietų Sąjungos rinktinės žaidėjų, kurie sužaidė bent po 50 oficialių rungtynių Lietuvos rinktinėje.
\paragraph{250\,000 steps, R-1=0.136, R-2=0, R-L=0.136}
Lietuvos krepšinio federacijos (LKF) generalinis sekretorius Mindaugas Špokas pirmadienį socialiniame tinkle „Facebook“ pasidalino statistika, kurios viršūnėje yra Arvydas Sabonis. Tai pats aukščiausias rezultatyvumo vidurkis tarp visų sovietų komandai atstovavusių žaidėjų.
\paragraph{300\,000 steps, R-1=0.175, R-2=0.036, R-L=0.175}
Lietuvos krepšinio legenda Arvydas Sabonis yra geriausias visų laikų rezultatyviausias krepšininkas tarp visų sovietų rinktinei atstovavusių žaidėjų, skaičiuojant tuos, kurie sužaidė ne mažiau nei 50 oficialių rungtynių.
\paragraph{350\,000 steps, R-1=0.364, R-2=0.038, R-L=0.255}
Buvęs Lietuvos krepšinio rinktinės vidurio puolėjas Valdis Valteris ir buvęs Kauno „Žalgirio“ krepšininkas Rimas Kurtinaitis yra tarp rezultatyviausių visų laikų Lietuvos rinktinės žaidėjų.

\end{document}